\def\year{2019}\relax
\documentclass[letterpaper]{article} 
\usepackage{aaai19}  
\usepackage{times}  
\usepackage{helvet}  
\usepackage{color}
\usepackage{courier}  
\usepackage{url}  
\usepackage{graphicx}  
\usepackage{mathtools}
\usepackage{amsfonts}
\usepackage{amsmath}
\usepackage{booktabs}
\usepackage{hyperref}

\DeclarePairedDelimiter\norm{\lVert}{\rVert}
\makeatletter
\renewcommand\@biblabel[1]{}
\makeatother

\begin{document}
%
\title{Knowledge Transfer
via Distillation of Activation Boundaries \\ Formed by Hidden Neurons
}


\author{
  Byeongho Heo\textsuperscript{1}
  \,\,
  Minsik Lee\textsuperscript{2}
  \,\,
  Sangdoo Yun\textsuperscript{3}
  \,\,
  Jin Young Choi\textsuperscript{1}
  \\
  \small{\texttt{\{bhheo, jychoi\}@snu.ac.kr},\;
  \texttt{mleepaper@hanyang.ac.kr},\;
  \texttt{sangdoo.yun@navercorp.com}}\\
  \textsuperscript{1}Department of ECE, ASRI, Seoul National University, Korea\\
  \textsuperscript{2}Division of EE, Hanyang University, Korea\\
  \textsuperscript{3}Clova AI Research, NAVER Corp, Korea\\
}

\nocopyright
\maketitle
\begin{abstract}
An activation boundary for a neuron refers to a separating hyperplane that determines whether the neuron is activated or deactivated.
It has been long considered in neural networks that the activations of neurons, rather than their exact output values, play the most important role in forming classification-friendly partitions of the hidden feature space.
However, as far as we know, this aspect of neural networks has not been considered in the literature of knowledge transfer.
In this paper, we propose a knowledge transfer method via distillation of activation boundaries formed by hidden neurons.
For the distillation, we propose an activation transfer loss that has the minimum value when the boundaries generated by the student coincide with those by the teacher.
Since the activation transfer loss is not differentiable, we design a piecewise differentiable loss approximating the activation transfer loss.
By the proposed method, the student learns a separating boundary between activation region and deactivation region formed by each neuron in the teacher.
Through the experiments in various aspects of knowledge transfer,
it is verified that the proposed method outperforms the current state-of-the-art.
\footnote[1]{Code is available at \href{https://github.com/bhheo/AB_distillation}{\url{https://github.com/bhheo/AB_distillation}}}
\end{abstract}

\section{Introduction}

Learning deep neural networks requires much time and computational resource.
Many studies have been conducted to make the learning process lighter and faster, one of which is to transfer knowledge.
Knowledge transfer is to accelerate and improve learning of a new network (student) by transferring knowledge from a previously learned network (teacher).
If the network architectures are identical, knowledge transfer can be simply conducted by copying the parameters from teacher to student.
However, in general, knowledge transfer aim to improve the performance of a small network (student) based on a large network (teacher).
Knowledge transfer between different network architectures is a challenging problem and has not been completely solved.

The main direction of knowledge transfer is to transfer neuron responses of hidden layers.
\citeauthor{FITNET}~\shortcite{FITNET} proposed a knowledge transfer method based on hidden layer responses.
\citeauthor{Attention}~\shortcite{Attention} changed the form of hidden neuron responses to an attention map, while \citeauthor{Gift_distill}~\shortcite{Gift_distill} used the form of Gram matrix~\cite{Grammatrix}.
The hidden layers of a neural network contains a lot of information and is suitable for knowledge transfer.
However, due to the high dimensionality and non-linearity of the hidden layer neurons, it is not easy to achieve a perfect transfer.

The activation boundary is a separating hyperplane that determines whether neurons are active or deactivated.
In neural networks, the activation of neurons has been considered to be important for a long time.
Recently, regarding the activation boundary,
\citeauthor{linear_region}~\shortcite{linear_region} and \citeauthor{inference_region}~\shortcite{inference_region} reported that a neural network expresses a complex function with a combination of activation boundaries.
\citeauthor{expressiveness}~\shortcite{expressiveness} provided the fact that the decision boundary of a neural network is composed of a combination of activation boundaries.
These studies give an insight that the transfer of activation boundary information in the teacher to the student could greatly contribute to an improvement of knowledge transfer performance in classification problems, because a classification problem highly depends on the formation of decision boundaries among classes.

Based on this inspiration, in this paper, we propose a knowledge transfer method that focuses on the transfer of activation boundaries.
In contrast to the conventional methods focusing on the magnitudes of neuron responses,
the proposed method aims to transfer whether neurons are activated or not.
First, we propose an activation transfer loss that minimizes the difference of neuron activations between the teacher and the student.
The activation transfer loss does not consider the magnitude of a neuron response but considers whether the neuron is activated or not.
Because of this characteristic, the activation transfer loss is suitable for transferring the activation boundaries, but is not differentiable.
Thus, it cannot be minimized by the gradient descent method.
To obtain a differentiable loss having the same goal, we propose an alternative loss similar to the hinge loss of SVM~\cite{SVM}.
The alternative loss is an approximation of the activation transfer loss which can be minimized with gradient descent.
Through the alternative loss, the student network can learn the activation boundaries.

The proposed method is verified through various experiments.
First, the proposed method is compared with the state-of-the-art algorithms in various aspects of knowledge transfer.
Second, we extend experiments to transfer learning and compare the proposed method in more general case.
Finally, the proposed method is analyzed through several experiments.

\begin{figure*}[t]
\begin{center}
\centerline{\includegraphics[width=1.0\linewidth]{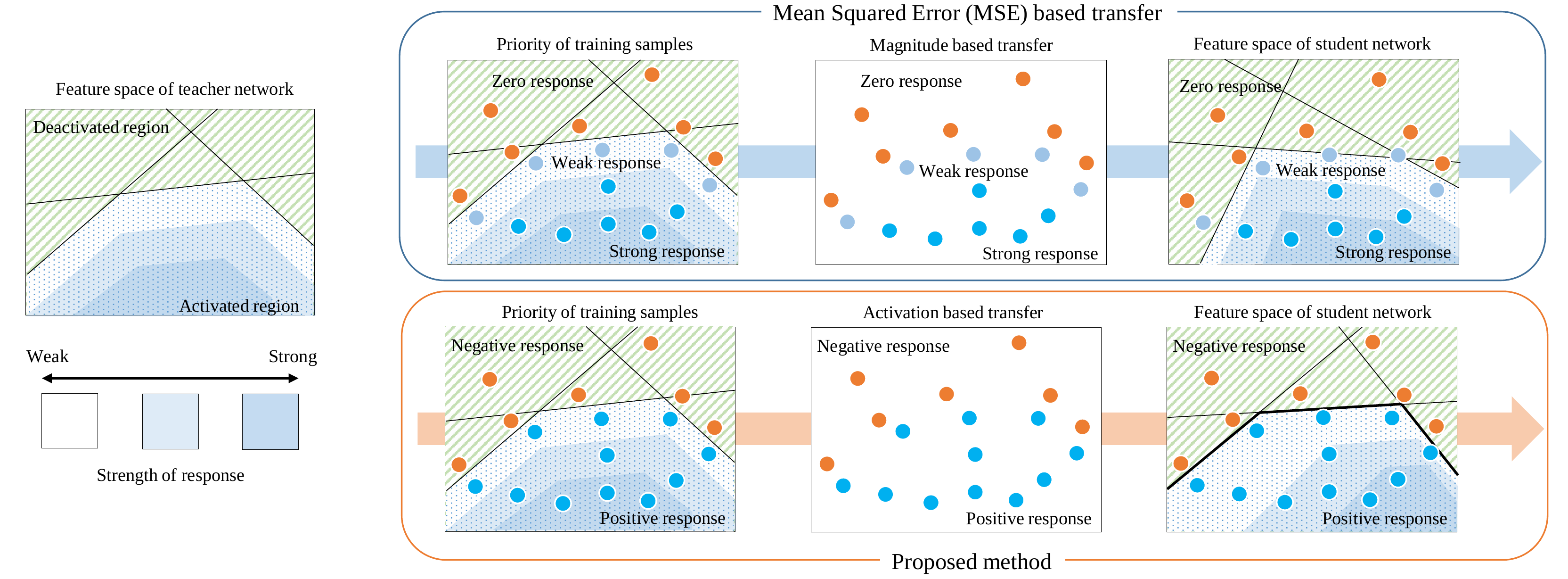}}
\caption{The concept of the proposed knowledge transfer method. The proposed method concentrates on the activation of neurons, not the magnitude of neuron responses. This concentration enables more precise transfer of the activation boundaries.}
\label{figure1}
\end{center}
\end{figure*}

\section{Related Work}

Rectified Linear Unit (ReLU)~\cite{relu} activation, which is less affected by the vanishing gradient problem, has become the basis of various network architectures~\cite{alexnet,googlenet,resnet}.
At the same time, studies on ReLU-based networks have been conducted.
ReLU activation divides the input space into two, and the dividing hyperplane between the two is called the activation boundary.
Activation boundaries of multiple neurons divide the input space several times.
Thus, the function expressed by the neural network is a set of piecewise linear partitions that are divided by the activation boundaries.
This is the appearance of a ReLU-based network that is commonly described in studies.~\cite{linear_region,inference_region,expressiveness}.
Based on this piecewise linear partitions, \citeauthor{linear_region}~\shortcite{linear_region} and \citeauthor{inference_region}~\shortcite{inference_region} propose to use the number of partitions divided by the activation boundaries as a measure of complexity of a network.
In other words, the complexity of a neural network depends on the placement of activation boundaries.
\citeauthor{expressiveness}~\shortcite{expressiveness} shows that the binary decision of a ReLU-based network is the same as the combination of threshold functions at the activation boundaries.
This means that the knowledge of a ReLU-based network is concentrated at the activation boundaries.
These studies imply that an activation boundary is a key factor of ReLU-based networks.
Despite such importance of activation boundaries, conventional knowledge transfer methods, as far as we know, do not aim to transfer activation boundaries.

\citeauthor{FITNET}~\shortcite{FITNET} first proposed a method of transferring the neuron responses of hidden layers for knowledge distillation.
In this method, the entire neurons in hidden layers were transferred using the mean square error function.
This approach was effective for the high-level hidden layers but not for the low-level hidden layers.
Subsequent studies have attempted to reduce the dimension of neurons to cope with low-level hidden layers.
In a convolutional neural network, the hidden layer response is a tensor consisting of two spatial dimensions and one channel dimension.
\citeauthor{Gift_distill}~\shortcite{Gift_distill} focus on channel dimensions and reduce spatial information.
In this work, only correlation between channels is transferred using a method similar to Gram matrix~\cite{Grammatrix}.
Conversely, \citeauthor{Attention}~\shortcite{Attention} concentrates on spatial attention and reduces channel information.
Neuron responses are averaged over the channel and only spatial information is transferred.
Both method succeed in transferring low-level hidden layers, however, both channel and spatial information are important information, and reducing dimension is a loss of information.
If low-level hidden layers can be transferred without dimension reduction, the performance of knowledge transfer might be improved.
Therefore, using activation boundaries, we challenge to transfer the entire neuron responses without reducing dimension.

\section{Method}

There are two types of learning schemes for transferring the responses of the hidden layer neurons from a teacher network to a student network.
The first approach transfers the neuron responses during the student neural network learns to classify input images~\cite{Attention,Jacobian,Hinton}.
In this case, the cross entropy loss and the transfer loss are combined to form a single integrated loss which is used for the whole training process.
The second approach is to transfer the neuron responses for initializing the student network before the classification training~\cite{FITNET,Gift_distill}.
After the network is initialized based on the transfer loss, it is trained for classification based on the cross-entropy loss.
In this paper, we discuss knowledge transfer based on the second approach, which is the network initialization method, to analyze the transfer loss independently.

A part of the teacher network, the part from the input image to a hidden layer, is defined as a function $\mathcal{T}$, and a part of the student network is defined as a function $\mathcal{S}$.
For an image $\boldsymbol{I}$, the neuron response vector of the hidden layer is $\mathcal{T}(\boldsymbol{I}) \in \mathbb{R}^M$ for the teacher and $\mathcal{S}(\boldsymbol{I}) \in \mathbb{R}^M$ for the student.
$M$ is the number of neurons in the hidden layer.
For convenience, we assume that the teacher and the student have hidden layers of the same size.
In order to describe the activation of neurons, $\mathcal{T}(\boldsymbol{I})$ and $\mathcal{S}(\boldsymbol{I})$ are defined as the values before the activation functions.
Here, we consider the element-wise ReLU $\sigma(x) = max(0,x)$ as the activation function.

An existing method~\cite{FITNET} of neuron response transfer is as follows.
\begin{equation}
\mathcal{L} (\boldsymbol{I}) = \norm{{\sigma(\mathcal{T}(\boldsymbol{I})) - \sigma(\mathcal{S}(\boldsymbol{I}))}}^2_2
\label{L2loss}
\end{equation}
It is the mean square error function between neuron responses after the ReLU activation.
(\ref{L2loss}) makes a student merely approximate the neuron responses of the teacher, but the resulting activation boundaries can be wildly different.
Figure~\ref{figure1} shows an example.
The mean square error is a loss that is biased to a large difference, so this will mainly transfer samples with strong responses.
However, as shown in Figure~\ref{figure1}, the activation boundaries are between weak responses and zero responses.
The mean square error can hardly distinguish between weak responses and zero responses, so it is difficult to transfer the activation boundaries by the method based on (~\ref{L2loss}).

As reported in \citeauthor{expressiveness}~\shortcite{expressiveness}, the activation boundaries play an important role in forming the decision boundaries for classification-friendly partitioning of the feature space in each hidden layer.
Thus it is essential to transfer  the activation boundaries accurately for knowledge transfer.
To transfer the activation boundaries accurately, our idea is to amplify the negligible transfer loss at the region near the activation boundaries.
To amplify the loss, we define
an element-wise activation indicator function that expresses whether a neuron is activated or not, i.e.,
\begin{equation}
\rho(x) =
\begin{cases}
1, & \text{if } x > 0\\
0, & \text{otherwise}.
\end{cases}
\label{indicator}
\end{equation}
The new loss to transfer the neuron activation is given by
\begin{equation}
\mathcal{L} (\boldsymbol{I}) = \norm{{\rho(\mathcal{T}(\boldsymbol{I})) - \rho(\mathcal{S}(\boldsymbol{I}))}}_1,
\label{binary}
\end{equation}
which is referred to as the {\it activation transfer loss} in our paper.
The activation transfer loss is a function that gives a constant penalty when the activations of teacher and student are different.
Regardless of the magnitude of neuron responses, all neurons have the same weight.
The lower part of Figure~\ref{figure1} shows knowledge transfer using the activation transfer loss.
Although the magnitude of neuron responses are not well transferred, it is trained to keep the activation of teacher neurons.
Thus, the activation boundaries are accurately transferred.
Considering the importance of activation boundaries in a neural network, the activation transfer loss is more efficient in knowledge transfer than the mean square error.

The activation transfer loss is suitable for transfer of activation boundaries.
However, since $\rho()$ is a discrete function, the activation transfer loss can not be minimized by gradient descent.
Therefore, we propose an alternative loss that can be minimized by gradient descent.
Minimizing the activation transfer loss is similar to learning a binary classifier.
The activation of teacher neurons $\rho(\mathcal{T}(\boldsymbol{I}))$ correspond to class labels.
The response of student neuron should be greater than 0 if the teacher neuron is active and less than 0 if the teacher neuron is deactivated.
Inspired by this similarity, the alternative loss is designed similar to the hinge loss~\cite{hinge} used in SVM~\cite{SVM}.
\begin{align}
\mathcal{L} (\boldsymbol{I}) = \norm{&\rho(\mathcal{T}(\boldsymbol{I})) \odot \sigma(\mu\boldsymbol{1} - \mathcal{S}(\boldsymbol{I})) \nonumber \\
& + (\boldsymbol{1} - \rho(\mathcal{T}(\boldsymbol{I}))) \odot \sigma(\mu\boldsymbol{1} + \mathcal{S}(\boldsymbol{I}))}_2^2
\label{alternative}
\end{align}
The symbol $\odot$ means element-wise product of vectors.
The $\boldsymbol{1}$ represents a vector of length $M$ where all component values are equal to one.
The alternative loss gives a square penalty for student neurons with different activations, and does not care about neurons with the same activation.
In addition, the margin $\mu$ is introduced for stability of training.
To understand the alternative loss function, we explain it in terms of gradients.
For convenience, the $i$-th element of neuron response vector is denoted as $t_i$ (teacher) and $s_i$ (student).
The derivative of (\ref{alternative}) with respect to $s_i$ is
\begin{equation}
- \frac{\partial \mathcal{L} (\boldsymbol{I})} {\partial s_i} =
\begin{cases}
2(s_i - \mu) ,& \text{if } \rho(t_i) = 1 \text{ and } s_i < \mu \\
-2(s_i + \mu) ,& \text{if } \rho(t_i) = 0 \text{ and } s_i > -\mu \\
0 ,& \text{otherwise}.
\end{cases}
\label{derivative}
\end{equation}
When the teacher neuron is activated and the student response is less than a margin $\mu$, then the gradient becomes positive.
Conversely, when the teacher neuron is deactivated and the student response is greater than the minus margin $-\mu$, the gradient becomes negative.
Otherwise, the gradient is zero.
Therefore, minimizing the alternative loss makes neurons of the teacher and student have the same activation, and puts a margin to the student's neuron response for stability.
The proposed method is to transfer neuron responses using the alternative loss.
Since the alternative loss approximates the activation transfer loss, the proposed method transfers the activation boundaries.
Effectiveness of the proposed method and the fact that the alternative loss actually approximates the activation transfer loss are verified in the experiment section.

\begin{table*}[t]
\centering
\caption{Learning speed experiment using various training epochs. Table shows error rate(\%) on test set.}
\begin{tabular}{@{}rcccccc@{}}
\toprule
Training epochs (initialize + training) & 1+1              & 1+5              & 3+12            & 5+25            & 10+50           & 20+100          \\ \midrule
Cross-Entropy training               & 43.37\%          & 17.72\%          & 11.76\%         & 8.63\%          & 7.41\%          & 6.47\%          \\
Knowledge Distillation (KD) training & 48.42\%          & 19.80\%          & 12.09\%         & 8.66\%          & 6.80\%          & 6.19\%          \\ \midrule
FITNET~\cite{FITNET} + KD    		 & 48.16\%          & 19.82\%          & 11.10\%         & 8.38\%          & 7.02\%          & 6.28\%          \\
FSP~\cite{Gift_distill} + KD 		 & 43.51\%          & 19.29\%          & 11.15\%         & 8.48\%          & 6.87\%          & 6.22\%          \\
AT~\cite{Attention} + KD  			 & 37.66\%          & 14.14\%          & 8.35\%          & 6.68\%          & 5.94\%          & 5.80\%          \\
Jacobian~\cite{Jacobian} + KD		 & 38.46\%          & 14.29\%          & 8.37\%          & 6.98\%          & 5.98\%          & 5.77\%          \\
Proposed + KD                        & \textbf{16.39}\% & \textbf{11.16}\% & \textbf{6.95}\% & \textbf{6.08}\% & \textbf{5.72}\% & \textbf{5.58}\% \\ \bottomrule
\end{tabular}
\label{fast}
\end{table*}

\subsection{Number of neurons}

In the previous section, we explain only the case where the number of neurons in a teacher and a student are the same.
However, if architecture of a teacher and a student are different, such as in the case of network compression, the number of neurons might be different.
In this case, we use a connector function that consists of simple operations~\cite{FITNET}.
Let the number of neurons in a teacher network be $M$ ($\mathcal{T}(\boldsymbol{I}) \in \mathbb{R}^M$), and the number of neurons in a student network $N$ ($\mathcal{S}(\boldsymbol{I}) \in \mathbb{R}^N$).
The connector function $r:\mathbb{R}^N \rightarrow \mathbb{R}^M$ converts a neuron response vector of student to the size of teacher vector.
Using the connector function, the alternative loss (\ref{alternative}) is changed as
\begin{align}
\mathcal{L} (\boldsymbol{I}) = \norm{&\rho(\mathcal{T}(\boldsymbol{I})) \odot \sigma(\mu\boldsymbol{1} - r(\mathcal{S}(\boldsymbol{I}))) \nonumber \\
& + (\boldsymbol{1} - \rho(\mathcal{T}(\boldsymbol{I}))) \odot \sigma(\mu\boldsymbol{1} + r(\mathcal{S}(\boldsymbol{I})))}_2^2.
\label{connect}
\end{align}
Typically, the connector function is a fully connected layer or a combination of a fully connected layer and a batch normalization layer.
In the initialization process, the connector function $r$ and the student network $\mathcal{S}$ are trained simultaneously to minimize (\ref{connect}).
After the initialization for knowledge transfer, only the student network without the connector function is used for classification.
\citeauthor{FITNET} first propose the use of connector functions in knowledge transfer.
It is based on an idea that if a student can represent a teacher with a simple converting operation, the student has the same level of knowledge as the teacher.
The connector function is effectively applied to knowledge transfer and allows the proposed method to be applied even when the number of neurons is different.

\subsection{Convolutional neural network}

Convolutional neural networks take an image as input and create hidden neuron responses with spatial dimensions.
Let the number of hidden neurons be $H \times W \times M$ (height, width, channel).
Since a convolutional operation shares weights for each spatial location, $H \times W \times M$ neuron responses can be interpreted as an $M$ dimensional neuron response for $H \times W$ receptive fields on an image~\cite{LeNet,FCN}.
Therefore, the proposed method can be applied to each spatial location of the neuron responses.
For a neuron response tensor $\mathcal{T}(\boldsymbol{I}) \in \mathbb{R}^{H \times W \times M}$ and $\mathcal{S}(\boldsymbol{I}) \in \mathbb{R}^{H \times W \times N}$,
let the neuron response vector at the ($i, j$)-th spatial position be $\mathcal{T}(\boldsymbol{I})_{ij} \in \mathbb{R}^M$ and $\mathcal{S}(\boldsymbol{I})_{ij} \in \mathbb{R}^N, (i = 1, \cdots ,H; j = 1, \cdots, W)$, respectively.
The proposed loss is represented as a sum over spatial locations.
\begin{align}
\mathcal{L} (\boldsymbol{I}) &= \sum_i^H \sum_j^W \norm{\rho(\mathcal{T}(\boldsymbol{I})_{ij}) \odot \sigma(\mu\boldsymbol{1} - r(\mathcal{S}(\boldsymbol{I})_{ij})) \nonumber \\
& + (\boldsymbol{1} - \rho(\mathcal{T}(\boldsymbol{I})_{ij})) \odot \sigma(\mu\boldsymbol{1} + r(\mathcal{S}(\boldsymbol{I})_{ij}))}_2^2.
\label{cnn}
\end{align}
Using (\ref{cnn}), the proposed method can also be applied to a convolutional neural network.
In this case, the connector function $r$ is shared at all spatial locations, so a 1$\times$1 convolutional layer is used instead of a fully connected layer.
In this case, teacher and student layers are forced to be of the same spatial size.
Therefore, like other knowledge transfer methods~\cite{FITNET,Attention,Jacobian}, a hidden layer which has the same spatial size as that of the teacher is selected for the target of knowledge transfer.

\section{Experiments}

We verified the proposed method through various experiments.
The first experiment is about various aspects of knowledge transfer.
Experiments that can verify the benefits of knowledge transfer are designed.
Through experiments, we show that the proposed method outperforms the current state-of-the-art in most aspects.
For the next experiment, we evaluated the proposed method in transfer learning to verify its performance in more general configurations.
In transfer learning, using a teacher pre-trained on ImageNet~\cite{ImageNet}, a randomly initialized student was trained on a target dataset other than ImageNet.
We measured performance on the scene classification dataset~\cite{MIT} and the fine-grained bird classification dataset~\cite{CUB} that have different properties from ImageNet.
Finally, several analytical experiments were conducted to help understand the proposed method.

\begin{table*}[t]
\caption{Performance for small size training data. Table shows error rate (\%) on test set.}
\centering
\begin{tabular}{@{}rcccccc@{}}
\toprule
Percentage of training data          & 0.1\%            & 1\%              & 5\%              & 10\%             & 20\%            & 100\%           \\ \midrule
Cross-Entropy training               & 73.83\%          & 48.41\%          & 28.12\%          & 21.76\%          & 15.26\%         & 6.47\%          \\
Knowledge Distillation (KD) training & 74.62\%          & 48.34\%          & 28.13\%          & 21.21\%          & 14.62\%         & 6.19\%          \\ \midrule
FITNET~\cite{FITNET} + KD            & 74.81\%          & 49.91\%          & 27.28\%          & 21.42\%          & 14.52\%         & 6.28\%          \\
FSP~\cite{Gift_distill} + KD         & 73.96\%          & 47.98\%          & 26.90\%          & 20.80\%          & 13.85\%         & 6.22\%          \\
AT~\cite{Attention} + KD             & 67.54\%          & 37.11\%          & 18.86\%          & 15.61\%          & 9.94\%          & 5.80\%          \\
Jacobian~\cite{Jacobian} + KD        & 68.65\%          & 36.99\%          & 18.34\%          & 15.03\%          & 9.83\%          & 5.77\%          \\
Proposed + KD                        & \textbf{50.32}\% & \textbf{21.54}\% & \textbf{14.99}\% & \textbf{13.09}\% & \textbf{9.16}\% & \textbf{5.58}\% \\ \bottomrule
\end{tabular}
\label{small}
\end{table*}

\begin{table*}[t]
\caption{Performance for various sizes of networks. Table shows error rate(\%) on test set.}
\centering
\begin{tabular}{@{}cccc|cccccc@{}}
\toprule
Compression type & Teacher  & Student  & Size ratio       & KD      & FITNET  & FSP     & AT      & Jacobian & Proposed         \\ \midrule
Depth            & WRN 22-4 & WRN 10-4 & 27.9\%           & 22.98\% & 23.34\% & 22.99\% & 18.06\% & 18.28\%  & \textbf{14.05}\% \\
Channel          & WRN 16-4 & WRN 16-2 & 25.2\%           & 20.48\% & 19.98\% & 19.78\% & 14.81\% & 14.41\%  & \textbf{11.62}\% \\
Depth \& Channel & WRN 22-4 & WRN 16-2 & 16.1\%           & 21.21\% & 21.42\% & 20.80\% & 15.61\% & 15.03\%  & \textbf{13.09}\% \\
Tiny network     & WRN 22-4 & WRN 10-1 & 1.8\%            & 29.57\% & 29.18\% & 28.70\% & 29.44\% & 28.70\%  & \textbf{23.27}\% \\
Same network     & WRN 16-4 & WRN 16-4 & 100\%            & 18.29\% & 17.91\% & 17.81\% & 12.03\% & 11.28\%  & \textbf{6.63}\%  \\ \bottomrule
\end{tabular}
\label{type}
\end{table*}

\subsection{Knowledge transfer for the same task}

In this section, experiments were performed on the CIFAR-10~\cite{CIFAR} dataset.
Target datasets of the teacher and the student are the CIFAR-10 dataset.
Wide residual networks (WRN)~\cite{WRN} which can change depth and the number of channels were used as the base network architecture.
A specific type of WRN is denoted as WRN$x$-$y$, where $x$ is the number of layers in the network, ie. depth, and $y$ is the multiplication factor of the channels.
For example, WRN22-4 is composed of 22 layers and is using 4 times the number of base channels.
As baseline, we used the Knowledge Distillation (KD) loss~\cite{Hinton} which is commonly used for knowledge transfer in the same dataset.
We have also presented the training result without any knowledge transfer.
The compared algorithms were implemented based on the author-provided codes and the papers.
Details on implementation are discussed in the last subsection.

First, we compared the algorithms in terms of learning speed.
Since knowledge transfer uses information of a pre-trained network, it can accelerate the training of a new network.
The learning speed was measured in terms of classification performance on various training epochs.
Experiments were conducted with WRN22-4 as a teacher network and WRN16-2 as a student network.
In this case, the error rate of the teacher is 4.51\%.
Table~\ref{fast} shows the results.
The top row of Table 1 indicates the numbers of epochs used for network initialization and classification training.
The rest of the rows represent the classification error (\%) of the student network on the test set.
Most knowledge transfer algorithms accelerate the learning of the student network.
Therefore, as the training epoch decreases, their relative performance improvement increases.
Here, we can confirm that the proposed method outperforms other algorithms in all cases.
In particular, the performance gap becomes larger as the training epochs becomes shorter.
Therefore, the proposed method is superior to the current state-of-the-art in terms of learning speed.

The second experiment is about small training data.
When the size of training data is small, it is difficult to achieve good generalization performance.
If there is a teacher network with a good generalization performance, knowledge transfer can help the student network to improve its generalization performance.
In order to test this property, we measured the performance when the training data was limited to a certain ratio.
As in the previous experiment, WRN22-4 and WRN16-2 were used as the teacher and the student, respectively.
The teacher network was trained using the whole training data, and the error rate was 4.51\%.
For training the student network, if the ratio of training data was more than 10\%, the number of training epochs was fixed to 120 epochs.
If the ratio was less than 10\%, we increased the number of epochs to maintain a certain amount of training iterations.
240 epochs were used for 5\%, 1,200 epochs for 1\%, and 12,000 epochs were used for 0.1\%.
The results are shown in Table~\ref{small}.
Most knowledge transfer methods help the student classifier generalize well, and the smaller the learning data, the greater the performance improvement by knowledge transfer.
Again in this case, the proposed method shows the best performance for all scenarios.
Especially, the proposed method is more effective when there is little training data.
Therefore, the proposed method can be the most effective choice for a small amount of training data.

\begin{table*}[t]
\caption{Performance of transfer learning on MIT scenes. Table shows accuracy (\%) on test set.}
\centering
\begin{tabular}{@{}rcccccc@{}}
\toprule
Number of training data per class  & 5                & 10               & 20               & 40               & 80               \\ \midrule
ImageNet pre-trained network & 32.39\%          & 42.46\%          & 52.99\%          & 62.54\%          & 69.78\%          \\ \midrule
randomly initialized network            & 11.12\%          & 17.99\%          & 30.37\%          & 43.88\%          & 56.94\%          \\
DTL on randomly initialized network  & 20.52\%          & 34.55\%          & 49.78\%          & 59.48\%          & 64.93\%          \\ \midrule
FITNET~\cite{FITNET} + DTL              & 29.55\%          & 43.81\%          & 53.51\%          & 63.36\%          & 66.57\%          \\
FSP~\cite{Gift_distill} + DTL           & 25.60\%          & 39.63\%          & 51.42\%          & 60.08\%          & 63.73\%          \\
AT~\cite{Attention} + DTL               & 22.61\%          & 35.37\%          & 49.10\%          & 60.15\%          & 65.15\%          \\
Jacobian~\cite{Jacobian} + DTL          & 26.64\%          & 38.28\%          & 51.34\%          & 61.19\%          & 63.96\%          \\
Proposed + DTL               & \textbf{43.36}\% & \textbf{56.72}\% & \textbf{66.34}\% & \textbf{70.75}\% & \textbf{74.10}\% \\ \bottomrule

\end{tabular}
\label{MIT}
\end{table*}

\begin{table*}[t]
\caption{Performance of transfer learning on CUB 2011 dataset. Table shows accuracy (\%) on test set.}
\centering
\begin{tabular}{@{}rcccccc@{}}
\toprule
Number of training data per class 		& 1                & 5                & 10               & 20               & 30               \\ \midrule
ImageNet pre-trained network 			& 8.01\%           & 29.84\%          & 48.83\%          & 69.02\%          & 74.70\%          \\ \midrule
randomly initialized network  			& 2.40\%           & 10.68\%          & 25.54\%          & 44.43\%          & 56.63\%          \\
DTL on randomly initialized network     & 5.35\%           & 26.29\%          & 42.15\%          & 48.24\%          & 62.05\%          \\ \midrule
FITNET~\cite{FITNET} + DTL              & 10.70\%          & 36.75\%          & 50.74\%          & 52.85\%          & 63.93\%          \\
FSP~\cite{Gift_distill} + DTL           & 7.06\%           & 29.43\%          & 42.44\%          & 43.96\%          & 56.52\%          \\
AT~\cite{Attention} + DTL               & 6.65\%           & 28.56\%          & 44.43\%          & 52.80\%          & 63.70\%          \\
Jacobian~\cite{Jacobian} + DTL          & 7.85\%           & 31.02\%          & 45.39\%          & 49.41\%          & 60.11\%          \\
Proposed + DTL              			& \textbf{13.38}\% & \textbf{45.39}\%  & \textbf{57.11}\%  & \textbf{62.93}\%  & \textbf{70.54}\%  \\ \bottomrule
\end{tabular}
\label{CUB}
\end{table*}

The last experiment is about knowledge transfer for networks of various sizes.
One of the main uses of knowledge transfer is for network compression and it has the advantage of being able to cope with various network sizes.
In this experiment, we measured the performance of knowledge transfer for various configurations of network sizes.
In this experiment, training was performed for 120 epochs using 10\% of the training data.
The experimental results are shown in Table~\ref{type}.
Here, we can see that all knowledge transfer algorithms, including the proposed method, are more vulnerable to depth compression than to channel compression.
However, the proposed method still achieves the best performance for all cases.
In particular, in case of ``Tiny network'', the proposed method greatly increases performance from the baseline, while the other methods have little improvement.
Therefore, the proposed method is better solution for the various type of network compression than other methods.

Through these three experiments, we have verified the proposed method in various configurations of knowledge transfer.
The proposed method can speed up the training process and helps the classifier generalize well.
In addition, it can cope with various kinds of network compression.
In all these aspects, the proposed method outperforms the current state-of-the-art.

\subsection{Transfer learning}
We conducted transfer learning to verify knowledge transfer in more general datasets.
When the training data of the target task is insufficient, pre-training a network in a large dataset, such as ImageNet~\cite{ImageNet}, is a great help for generalization performance.
However, pre-training on large datasets requires a lot of effort and computation.
The purpose of transfer learning is to achieve generalization performance without a separate pre-training on a large dataset.
Knowledge transfer in transfer learning uses an existing ImageNet-trained network as a teacher to train a student network in the target dataset. (That is, we don't want to pre-train the student on ImageNet explicitly.)
In this experiment, ResNet50~\cite{resnet} trained on ImageNet was used as a teacher network and Mobilenet~\cite{mobilenet} was used as a student network.
Mobilenet is one of the most widely used small networks and has a quite different structure than ResNet50.
Therefore, this is a very challenging case for knowledge transfer.
Recently, \citeauthor{Jacobian}~\shortcite{Jacobian} proposed a method to extend the KD loss~\cite{Hinton} to transfer learning.
We call this method distillation-in-transfer-learning (DTL).
Similar to the KD loss in the previous section, DTL was used as the baseline of transfer learning.

The MIT scenes dataset~\cite{MIT}, which is an indoor scene classification dataset, and the CUB 2011 dataset~\cite {CUB}, a fine-grained bird classification dataset, were used as target datasets of transfer learning.
In these two datasets, we experimented how knowledge transfer contributes to the generalization performance of classifiers.
The results for the MIT scenes dataset are shown in Table~\ref{MIT}, and Table~\ref{CUB} shows the performance for the CUB 2011 dataset.
The performance of the classifier is reported in terms of accuracy.
If a network is pre-trained on ImageNet, the classifier can achieve better performance than a randomly initialized network.
Here, we can see that all knowledge transfer methods, including the baseline, show significant improvement over the randomly initialized network.
However, most of them are not always better than a separate pre-training, and in fact, they are usually worse.
On the other hand, the proposed method achieves far better performance than all the other methods in all cases, and the performance improvement increases as the number of training samples becomes smaller.
In particular, the proposed method shows better performance than the ImageNet pre-trained network in majority of the cases.
Therefore, to achieve better generalization, the proposed method can be a better solution than the ImageNet pre-training, which is a big step in knowledge transfer.

\subsection{Analysis}

\begin{table}[t]
\centering
\caption{Layer-wise analysis of knowledge transfer. Table shows error rate (\%) of classification.}
\begin{tabular}{@{}rcccc@{}}
\toprule
                                     & Layer1           & Layer2           & Layer3           & Multi-layer      \\ \midrule
KD 			                         & \multicolumn{4}{c}{21.21\%}                            \\
FITNET + KD                          & 21.90\%          & 22.00\%          & 21.42\%          & 21.83\%          \\
FSP + KD                             & 21.55\%          & 21.28\%          & 21.27\%          & 20.80\%          \\
AT + KD                              & 20.18\%          & 17.64\%          & 17.48\%          & 15.61\%          \\
Jacobian + KD                        & 19.52\%          & 15.45\%          & 23.61\%          & 15.03\%          \\
Proposed + KD                        & \textbf{19.06}\% & \textbf{15.37}\% & \textbf{17.23}\% & \textbf{13.09}\% \\ \bottomrule
\end{tabular}
\label{layer}
\end{table}

\begin{table}[t]
\centering
\caption{Comparison of activation similarity with $l_p$ norm. Table shows percentage of same activation (\%) and classification error rate (\%).}
\begin{tabular}{@{}rcccc@{}}
\toprule
           			& $l_2$ loss & $l_1$ loss & $l_{0.5}$ loss & Proposed \\ \midrule
Layer 1     	& 56.1\%  & 66.9\%  & 68.8\%    & \textbf{96.3}\% \\
Layer 2     	& 67.3\%  & 72.5\%  & 73.0\%    & \textbf{96.4}\% \\
Layer 3    		& 56.0\%  & 56.1\%  & 56.4\%    & \textbf{92.8}\% \\ \midrule
Error rate		& 10.1\%  & 9.8\%  & 9.6\%    & \textbf{5.1}\% \\\bottomrule
\end{tabular}
\label{self}
\end{table}

\begin{table}[t]
\caption{Ablation study for margin value ($\mu$).}
\centering
\begin{tabular}{@{}rccccc@{}}
\toprule
Margin ($\mu$) & 0.75   & 1      & 1.5    & 2      & 4      \\ \midrule
Error rate   & 13.9\% & 13.1\% & 12.1\% & 12.1\% & 11.7\% \\ \bottomrule
\end{tabular}
\label{ablation}
\end{table}

We conducted additional experiments to help understand the proposed method.
These were knowledge distillation experiments using WRN on the CIFAR-10 dataset.
In WRN, residual layers are gathered to form one layer group, and three layer groups are gathered to form a network.
We name the last layer of each group as layer 1, layer 2, and layer 3.
The spatial sizes of these layers are 32$\times$32, 16$\times$16, and 8$\times$8, respectively.
Here, we regard these three layers as representatives for low-level, mid-level, and high-level layers, respectively.
We analyzed the characteristics of each layer in knowledge transfer through three analytical experiments.

The first experiment is about the effect of layer-wise transfer.
Knowledge transfer methods usually transfer multiple layer responses, but in this experiment, they are also performed for only a single layer to show performance differences according to hidden layer location.
Training was conducted for 120 epochs with 10\% of the training data.
The experimental results are reported in Table~\ref{layer}.
Most algorithms show the best performance when multiple layers are transferred, but oddly, FITNET shows the best performance when only layer 3 is transferred.
On the other hand, ``Jacobian'' has excellent performance for the mid-level layer but is not very good for the high-level layer.
Conversely, AT is good for the high-level layer, but is weak for the mid-level and low-level layers.
These results imply that a particular transfer method may not be good for all levels of layers.
Surprisingly, however, the proposed method shows the best performance in all configurations.
This indicates that the proposed method is indeed suitable for transferring general neuron responses, and explains the reason for the superb performance in the previous experiments.

The second experiment is about whether the alternative loss proposed in this paper makes the neuron activations similar.
For comparison, we also evaluated the $l_1$ and $l_{0.5}$ loss functions, in addition to the $l_2$ loss.
The experiment used all training data in CIFAR-10, and WRN16-4 was used for both the teacher and student networks.
Here, we used the same network for both the teacher and student to eliminate any interference caused by architectural differences.
The connector functions were not used for the same reason.
The experimental results are shown in Table~\ref{self}.
The percentages of neurons with the same activations in the teacher and student are presented here, and the classification errors are shown at the bottom.
Since a neuron's activation has two states, ideally, the similarity of those for a randomly initialized network should be 50\%.
Therefore, in this perspective, the $l_2$ loss is unsuccessful in transferring the activation of neurons.
Other $l_p$ losses make the activation more similar than the $l_2$ loss and also has better performance.
However, the proposed method has much higher similarity and much better performance.
This result implies that the proposed method is indeed a reasonable approximation of the ideal activation transfer loss.

The last is an ablation study on the margin value $\mu$, a parameter of the proposed method.
Although the margin value is arbitrarily set to one in other experiments, dependency of the proposed method on the margin value is an important issue.
Experiments were performed on CIFAR-10 with 10\% training data.
The experimental results are shown in Table~\ref{ablation}.
Here, we can see that when the margin is larger enough, there is not much a difference in performance.
The overall performance differences are not big, which means that the proposed method is not sensitive to the parameter.

\subsection{Implementation Details}
The compared algorithms were implemented based on the author-provided codes and the papers.
Based on the results of Table~\ref{layer}, all algorithms except FITNET were implemented to transfer multiple layer responses, and only FITNET was restricted to the last layer.
The last hidden layers among the layers with the same spatial size was selected as the targets of transfer.
Teacher and student layers with the same spatial size were matched for the knowledge transfer.
In the case of WRN on CIFAR-10, three layers were transferred, while for ResNet and Mobilenet, four layers with the same spatial size were used.
The relative weight for the transfer loss were tuned for each algorithm based on a weight search on a logarithmic grid.
The experiment using 10 \% training data in Table 2 and the experiment using 80 training data in Table~\ref{MIT} were the target experiment for the parameter search.
When the number of neurons is different, a combination of a convolution layer and a batch normalization layer was used as a connector function.
The connector function is applied to FSP and FITNET as well as the proposed method.
For the CIFAR-10 dataset, AT and ``Jacobian'' used only their corresponding transfer losses for the epochs in the initialization phase and their corresponding transfer losses in conjunction with the KD loss for the remaining epochs.
FITNET, FSP, and proposed method used only their corresponding transfer losses for initialization and use only the KD loss for the remaining epochs.
In transfer learning, training was conducted for 100 epochs.
For the methods requiring initialization, 60 epochs were used for initialization and 40 epochs were used for classification training.
AT and ``Jacobian'' were implemented to use the transfer loss, DTL loss, and cross-entropy loss simultaneously for 100 epochs.
The learning rate scheme for CIFAR-10 started at learning rate 0.1 and divided the learning rate by 5 at 30\%, 60\%, and 80\% of the total training epochs.
In transfer learning, the learning rate started at 0.01 and divided by 10 when the training was over half the total epochs.
In all cases, Nesterov momentum of 0.9 and a weight decay of $5\times 10^{-4}$ was used.

\section{Conclusion}

Activation boundaries play an important role in neural networks.
We have investigated the role of activation boundaries in knowledge transfer and concluded that transfer of activation boundaries can greatly enhance performance of knowledge transfer.
For this purpose, we have proposed a knowledge transfer method by transferring activation boundaries of hidden neurons.
Experiments have shown that the proposed method accurately transfers the activation boundaries.
At the same time, the proposed method shows much higher performance than the current state-of-the-art.
In particular, in many cases of transfer learning, the proposed method shows better performance than ImageNet pre-training, which is a great achievement in knowledge transfer.
In future, various tasks that originally relied on pre-training can be modified to accommodate knowledge transfer for better performance and learning speed.
The excellent performance of the proposed method implies that activation boundaries are indeed powerful for knowledge transfer, which can potentially change some of the essential practices of neural networks.

\section{Acknowledgement}

This work was supported by  
Next-Generation ICD Program through NRF funded by Ministry of S\&ICT [2017M3C4A7077582] and
ICT R\&D program of MSIP/IITP [No.B0101-15-0552, Predictive Visual Intelligence Technology].

\bibliographystyle{aaai}
\bibliography{refs}

\end{document}